\documentclass[conference]{IEEEtran}

\IEEEoverridecommandlockouts

\usepackage[numbers,sort&compress]{natbib}
\usepackage{url}

\usepackage{graphicx}
\usepackage{color,hyperref}
\definecolor{blue}{rgb}{0.0,0.0,1.0}
\hypersetup{colorlinks,breaklinks,
            linkcolor=black,urlcolor=blue,
            anchorcolor=blue,citecolor=blue}
\usepackage[vlined, ruled, boxed]{algorithm2e}
\graphicspath{{/}} % Paths to the graph folder
\usepackage{subcaption}

\usepackage{multirow}
\usepackage{diagbox}

\usepackage{amssymb,amsmath,amsthm}
\usepackage{amsfonts}
\usepackage{array}

\usepackage[pagewise,displaymath]{lineno}
\usepackage{eurosym}
\usepackage{bbm}
\usepackage{bm}

\usepackage{algorithm2e}

\usepackage[utf8]{inputenc}
\usepackage[T1]{fontenc}

\usepackage[colorinlistoftodos]{todonotes}

\makeatletter
\newcommand{\linebreakand}{%
  \end{@IEEEauthorhalign}
  \hfill\mbox{}\par
  \mbox{}\hfill\begin{@IEEEauthorhalign}
}
\makeatother

\begin{document}

\title{Use digital twins to support fault diagnosis from system-level condition-monitoring data} 

\author{\IEEEauthorblockN{Killian Mc Court}
\IEEEauthorblockA{CentraleSup\'{e}lec, \\ Universit\'{e} Paris-Saclay, \\ Gif-sur-Yvette, 91190,~France}
\and
\IEEEauthorblockN{Xavier Mc Court}
\IEEEauthorblockA{CentraleSup\'{e}lec, \\ Universit\'{e} Paris-Saclay, \\ Gif-sur-Yvette, 91190,~France}
\and
\IEEEauthorblockN{Shijia Du}
\IEEEauthorblockA{Shenzhen Weimin Institute for \\ Reliability Systems Engineering, \\ Shenzhen, China} 
\linebreakand 
%\IEEEauthorblockN{Lama Itani}
%\IEEEauthorblockA{Mathworks France, \\ Meudon, France}
%\and
\IEEEauthorblockN{Zhiguo Zeng}
\IEEEauthorblockA{Chair on Risk and Resilience of Complex Systems, \\ Laboratoie Genie Industriel, CentraleSup\'{e}lec, \\ Universit\'{e} Paris-Saclay, 91190, Gif-sur-Yvette,~France}
}

\maketitle

% \todoInline{@all: Could you please check the name spelling and affiliations?}

%\pagestyle{empty}  % no page number for the second and the later pages

\begin{abstract}
	Deep learning models have created great opportunities for data-driven fault diagnosis but they requires large amount of labeled failure data for training. In this paper, we propose to use a digital twin to support developing data-driven fault diagnosis model to reduce the amount of failure data used in the training process. The developed fault diagnosis models are also able to diagnose component-level failures based on system-level condition-monitoring data. The proposed framework is evaluated on a real-world robot system. The results showed that the deep learning model trained by digital twins is able to diagnose the locations and modes of $9$ faults/failure from $4$ different motors. However, the performance of the model trained by a digital twin can still be improved, especially when the digital twin model has some discrepancy with the real system.
\end{abstract}



% Glossaries
% Print the glossary
%\setlength{\glsdescwidth}{.8\textwidth}
%\printglossary[type=\acronymtype,style=long]
%\printnomenclature[1.5cm]

% \IEEEpeerreviewmaketitle
%\thispagestyle{empty} % no page number for the first page

%\linenumbers

\section{Introduction}\label{Sect_Intro}

Fault diagnosis is an essential task in reliability and predictive maintenance \cite{zhong_overview_2023}. It collects and analyzes condition-monitoring data from sensors to diagnose the location and cause of failures \cite{zhu_survey_2024}. The rapid advancements in artificial intelligence (AI) have dramatically transformed fault diagnosis, with deep learning-based models becoming prevalent and showing great success in both academia and industry \cite{he_deep_2017}. For example, convolutional neural networks (CNNs) have been employed in automated fault detection for machinery vibrations \cite{xia_fault_2018}, while recurrent neural networks (RNNs) like LSTM have proven useful in diagnosing faults based on time series data \cite{shi_planetary_2022}. 

The deep learning-based fault diagnosis models, despite of their wide applications and great success, face notable limitations. First, they often require extensive amounts of labeled training data that are difficult to obtain in practice \cite{zhong_overview_2023}. Second, the majority of deep learning-based models rely on detailed, component-level monitoring data to accurately detect and localize component-level failure \cite{zhu_survey_2024}. For example, to detect and diagnose bearing failure, most existing models use bearing-level condition-monitoring signals like vibration, noise, etc \cite{xia_fault_2018}. In a large number of scenarios, however, condition-monitoring data can only be collected at the system-level, not the component-level, due to cost constraints, sensor limitations, or the physical inaccessibility of specific components.

In this paper, we attempt to address these two issues by leveraging the high-fidelity simulation and real-time updating capability of digital twins \cite{ge2023digital}. First, we present a new reference model of digital twin, called digital failure twin, that is specially designed for modeling and simulating failure behavior and support failure-related decision-making. Second, we demonstrate, through a real-world case study on a robot, how to use a digital failure twin to develop component-level fault diagnosis model from system-level condition-monitoring data. Finally, we created an open-source dataset that could serve as a benchmark for further research on digital twin-supported fault diagnosis and predictive maintenance studies.

\section{Motivating example}\label{Sect_2}

\begin{figure*}[htbp]
    \centering
    \includegraphics[width=.7\textwidth]{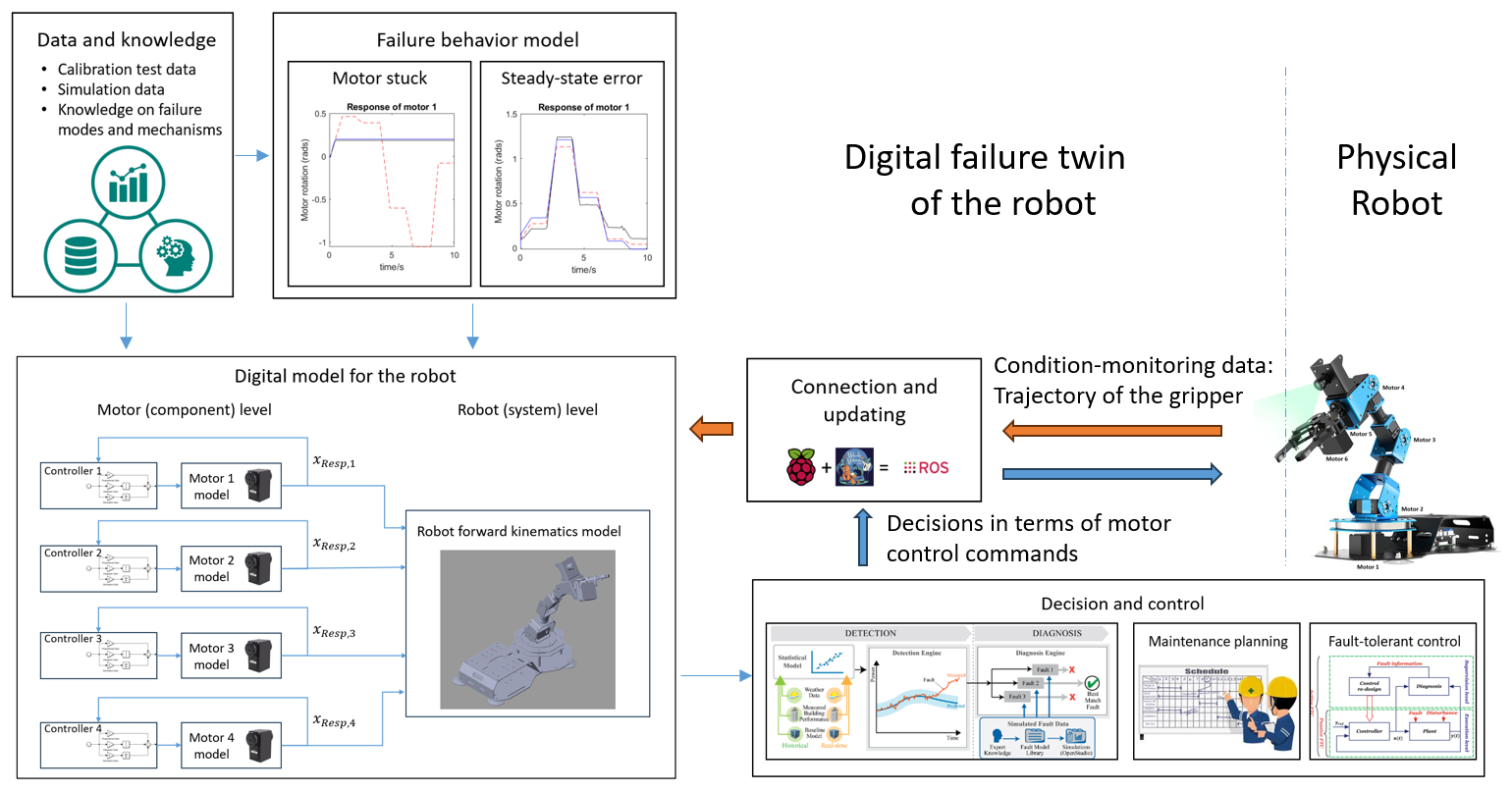}
    \caption{The developed digital failure twin for the robot.}
    \label{fig:dtrRobot}
\end{figure*}

As a motivating example, we consider an educational robot named ArmPi FPV from Hiwonder, as shown in Figure \ref{fig:dtrRobot}. This robot will be used as a case study to test the developed models in this paper. As can be seen from Figure \ref{fig:dtrRobot}, the robot consists of 5 joints and one end-effector (claw) and has six degrees of freedom. Each joint is controlled by a servo motor, and all the six servo motors can feedback their position, temperature and voltage on request. The robot is controlled by a Raspberry Pi $4$ mini-computer that runs an Ubuntu $18.04$ operating system. Robot Operating System (ROS) $1$ Melodic \cite{ROSMelodic} is used for managing the interaction and communication between the robot hardware and the software that controls it. The functional requirement of the robot is to control the end-effector to follow a given trajectory, under the specified performance requirements. For example, in order to collect an item, the end-effector has to follow the given trajectory command, and the performance requirement here is the accuracy and timeliness of the movement of the end-effector.

In this example, the component-level we considered is the motor-level, while the system-level condition-monitoring data are the movement trajectory of the end-effector. We consider three possible states for each motor: normal operation, stuck, and steady-state error failure. The motor $5$ and $6$ control the open/close and rotation of the end-effector, respectively, and do not affect the trajectory of the end-effector. Therefore, we only consider motors $1-4$ in this study. Let us introduce a new variable $y \in \{0, 1, 2, \cdots, 8\}$ where 
\begin{itemize}
    \item $y = 0$ represents that all the four motors are in normal operation states;
    \item $y = 1, 2, 3, 4$ represent that the motors $1, 2, 3, 4$ get stuck, respectively;
    \item $y = 5, 6, 7, 8$ represent that the motors $1, 2, 3, 4$ have steady-state error failure, respectively.
\end{itemize}
The variable $y$ therefore represents all the states we want to predict.

Let $\textbf{x}_k$ denote the system-level condition-monitoring data collected at time instant $k$, where $\textbf{x} \in \mathbb{R}^{d}$ and $d$ is the dimension of the condition-monitoring data. In this study, we have $d=7$: the commands to motors $1$-$4$ and the response trajectory of the end-effector ($x, y, z$ coordinates). In reality, the trajectory of the end-effector can be measured by placing an accelerometer on the end-effector and computing trajectory based on the measured acceleration of the three axes. The measured trajectory ($x, y, z$ coordinates), combined with the commands to the motors $1-4$, are the input data to the diagnosis algorithm. However, due to complexities of implementation, we did not use a real accelerometer to measure the trajectory of the end-effector. Rather, we use a virtual sensor to compute the trajectory through a forward kinematics model of the robot based on the measured position data of the four motors $1-4$ \cite{zhang2022robot}. It should be noted that the position data of the four motors are used only to compute the output of the virtual trajectory sensor, but not used as inputs to the fault diagnosis model. Therefore, using the virtual sensor, instead of a real one, does not affect the main objectives of our diagnosis framework.

\section{Digital failure twin for the robot}\label{Sect_3}

In this section, we present a digital failure twin for the robot to support its fault diagnosis, as shown in Figure \ref{fig:dtrRobot}. A digital failure twin is built from five essential modules, \textit{i.e.,} digital model, failure behavior model, data and knowledge, connection and updating, and decision and control. 

\subsection{Digital model}\label{Sect_DigitalModel}

The backbone of a digital failure twin is a digital model that simulates the performance under normal operation. The digital model of the robot is developed in Matlab Simulink and Simscape \cite{SimscapeMultibody}. A schematic of the developed digital model is shown in Figure \ref{fig:dtrRobot}. The objective of the digital model is to simulate the trajectory response of the end-effector, given the input commands on the four motors (component-level). Therefore, the inputs of the digital model are the commands on the four motors, denoted by $x_{cmd, 1}, x_{cmd, 2}, \cdots, x_{cmd, 4}$. The outputs of the digital model are the position response of the end-effector, denoted by the 3-dimensional coordinates of the end-effector $x, y$ and $z$. Both the input commands and the response trajectory can be a single value or a time series.

The digital model comprises of two levels. On the component level, it models the controllers and dynamic behavior of the four motors ($1$-$4$). The controllers for the motors are modeled as PID controllers in Simulink. The gains P, I and D are tuned based on experiment data. The motors' dynamics is modeled by the revolute joint block in Simscape Multibody \cite{SimscapeMultibody}, in which the motor behavior is approximated by a linear spring-damper model. Joint springs attempt to displace the joint primitive from its equilibrium position, and joint dampers act as energy dissipation elements. The damping coefficient of the damper and spring stiffness of the spring are two key parameters that determine the dynamic response of the motor that are tuned based on actual test results. The component-level models calculate the responses of the motors $x_{resp, 1}, x_{resp, 2}, x_{resp, 3}$ and $x_{resp, 4}$ from the input commands. The calculated responses are, then, fed into the system-level model to calculate the trajectory response of the end-effector. Here, we rely on a forward kinematics model to do the system-level simulation. The forward Kinematics model is developed based on a multibody tree using Simscape Multibody solver in MATLAB 2023b \cite{SimscapeMultibody}. 

\subsection{Failure behavior model}

As shown in Figure \ref{fig:dtrRobot}, a digital twin model should also be able to simulate the failure behavior. In this case study, we consider two failure modes, \emph{i.e.,} steady-state error and motor stuck for motors $1$ - $4$. The failure behavior model, then, focuses on simulating the behavior of these two failure modes. 

In practice, steady-state error occurs when after the response process stabilizes, the error between the steady-state response and the command exceeds its nominal value. It can be caused by the degradation of position sensors of the motor, leading to inadequate calculation of command/response difference, and therefore erroneous steady-state response. Let $\epsilon_{SS}$ represent the steady-state error. In this paper, we assume that when the distribution of $\epsilon_{SS}$ is
\begin{equation}\label{Eq_SteadyStateError}
    \epsilon_{SS} \sim \left\{
        \begin{aligned}
        U(\epsilon_l, \epsilon_u), & \text{ with probability } 0.5, \\
        -U(\epsilon_l, \epsilon_u), & \text{ with probability } 0.5.        
        \end{aligned}
    \right.
\end{equation} 
where $\epsilon_l$ and $\epsilon_u$ are the lower and upper bounds for the steady-state error when a failure occurs and need to be assigned by the modeller. Usually, we choose a value of $\epsilon$ that are significantly larger than the accuracy of the motor from its datasheet. Based on this assumption, the steady-state error failure can be simulated by generated a random noise and added to the control signals for each motor.

Motor stuck can be caused by factors like broken brushes, damaged power supply, etc. When a stuck failure occurs, a motor no longer rotates and its position freezes at the same value. To simulate this, we run a normal simulation to get the normal response and, then, manually fix the motor response to the value at the moment when the stuck occurs. 

\subsection{Other elements in the digital twin model}\label{Sect_OtherElements}

Figure \ref{fig:dtrRobot} shows a complete digital failure twin for the robot. Apart from the digital model and the failure behavior model, there are other three essential building blocks of this digital failure twin: data and knowledge, connection and updating, and decision and control. Data and knowledge block stores the data and knowledge needed for the digital twin model to function. Examples of data and knowledge used in this use case include the expert knowledge, historical data and calibration test data for calibrating the model parameters, knowledge regarding the motor failure mode and failure mechanisms, \emph{etc.} 

Connection and updating block is responsible for the communication between the digital twin model and the physical robot. In this case study, the data flow from the physical robot to its digital twin are mainly condition-monitoring data, \emph{i.e.,} the measured movement trajectory of the end-effector. The data flow on the other direction are mainly control commands on the motor level, reflecting the updated decision made by the decision and control block. Technically, the communication between the digital twin and the physical robot is realized through a Ros topic. For details, please refer to our open-sourced model and its documentation \cite{dtrRepo}. It should be noted that for this robot, condition-monitoring data on the motor level, \emph{e.g.,} its position, temperature and voltage can also be collected. However, they are not used in the diagnosis algorithm to be developed in the next Sections as the focus on this paper is to demonstrate the feasibility of diagnosing component-level failure based on only system-level condition-monitoring data. 

Decision and control block is responsible for solving decision-making problems based on the real-time data from the digital twin and generate updated control commands for the physical robot. In this paper, we only focus on a specific decision problem: Fault detection and diagnosis. Various other decision-making problems can also benefit from such a digital failure twin, \emph{e.g.,} maintenance planning \cite{mitici2023dynamic}, fault-tolerant control \cite{macgregor2012monitoring} and health-aware control \cite{pour2021health}.

\section{Fault diagnosis based on digital failure twin}\label{Sect_DTDevel}

\subsection{Generate training data from the digital failure twin}

\begin{figure}[htbp]
    \centering
    \includegraphics[width=.3\textwidth, height=.25\textwidth]{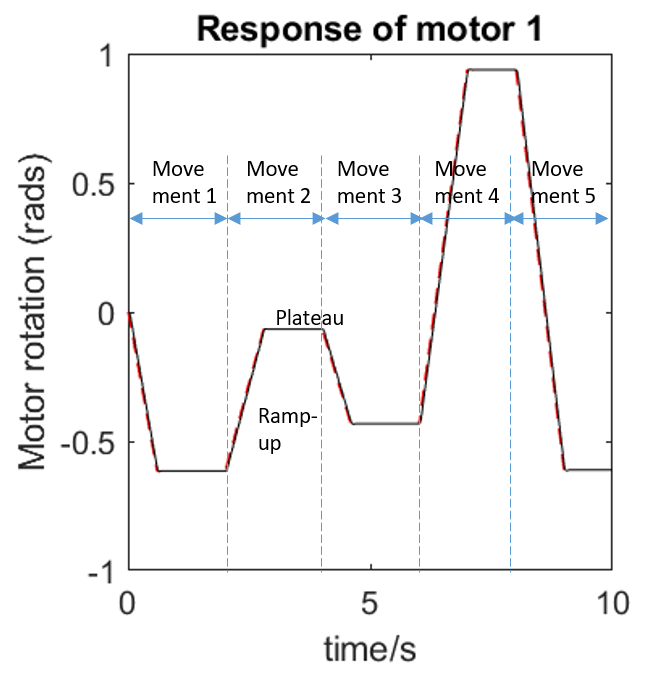}
    \caption{An illustration of one trajectory: Response of motor 1.}
    \label{fig:exampleTraj}
\end{figure}

We use the developed digital failure twin to generate training data for the fault diagnosis model. For each one of the nine labels, we generate $400$ trajectories. A sample trajectory is shown in Figure \ref{fig:exampleTraj}. In each trajectory, the robot perform five random movements within $10$ seconds. Each random movement takes two seconds and involve the movements of all the four motors. For a given motor, it moves to a random position within the range of the joint limit. The position is sampled from a uniform distribution. The motor movement further comprises of a ramp-up period, where the motor linearly turn to its destination, and a plateau period, where the motor holds on the current position until the end of this period. The ramp-up and plateau period together take $2$ seconds to complete.

For each trajectory, the commands on the four motors and the $x, y, z$ coordinates of the end-effector are saved as features for training the fault diagnosis model. The simulation is conducted in Matlab Simulink with a step size of $0.01$ seconds. Therefore, all the collected features are a time series of $1000$ points and all the features form a matrix of dimension $1000\times 7$. For $y=1, 2, \cdots, 8$, the corresponding failure mode is injected to generate the needed failure data. In this way, we can generate a total amount of $3600$ datasets with balanced labels among the nine classes.

\subsection{Collect test data from the real robot}

To evaluate the performance of the trained diagnosis model, an Armpi FPV robot is programmed to perform the same type of random movements as the training dataset (described in Sect. \ref{Sect_DTDevel}-A). For each label, we generate $10$ trajectories in the real robot so altogether we have $90$ trajectories from the real robot with balanced class labels. The test data are collected through Ros Melodic with a sampling frequency of $10$ Hz. The simulation step size used for creating the simulation dataset is $0.01$ seconds, which corresponds to a sampling frequency of $100$ Hz. Therefore, the original test data was enriched to $100$Hz through interpolation. The detailed procedures for creating and pre-processing the test data can be found in our online Github repository \cite{dtrRepo}.

\subsection{Training and evaluating the fault diagnosis model}

Since the input features are time series data, we chose to design a model architecture based on LSTM as LSTM is widely-acknowledged for being able to capture the time-dependencies in the time series data \cite{shi_planetary_2022}. The final model architecture is shown in Table \ref{table:layerDetails}. 

\begin{table}[htbp]
    \centering    
    \caption{Details of each layer.}
    \label{table:layerDetails}
    \begin{tabular}{ccp{3.8cm}}
        \hline
        Index & Type & Details \\
        \hline
        1 & Sequence input & Input size equals to the number of features.  \\
        2 & LSTM & Number of hidden units: $100$.  \\
        3 & Drop-out & Drop-out rate: $0.1$. \\
        4 & LSTM & Number of hidden units: $100$. \\
        5 & Fully connected & Output size: $9$. \\
        6 & Softmax & Softmax \\
        7 & Classification Output &  crossentropyex with $9$ classes.\\
        \hline       
    \end{tabular}
\end{table}

We first train a fault diagnosis model based on the digital twin-generated training dataset. The original features generated by the digital twin are used directly, \emph{i.e.,} the control commands on motors $1$-$4$ and the measured $x, y, z$ coordinates of the end-effector. The training dataset are randomly split into $90\%$ for training and $10\%$ for validation. The trained model is, then, applied to predict the labels of the test data collected from the real robot. The average accuracy over the $9$ classes, and the precision, recall and F1 score of each class are used as performance metrics to evaluate the performance of the model. All the experiments are repeated $5$ times to account for the randomness in the training process. 

In a second model, we first use the digital twin model to augment the original features. Instead of using the control commands on motors $1$-$4$ as features, we use these control commands to calculate the desired trajectory of the end-effector using the digital twin model (in terms of the $x, y, z$ coordinates). The desired trajectory is, then, compared to the actual measurements to calculate the residual errors on the $x, y$ and $z$ axes. Then, we use the desired trajectories ($x, y, z$) and the residual errors on the three axes as input features to train a fault diagnosis model. The size of input features in this case is $6\times 1000$. The other settings are kept the same as the first case to ensure a fair comparison.

The training is conducted using the Adam optimizer with a learning rate of $.0001$ and a batch size of $32$. After some initial trials, training is terminated after $5000$ epochs to achieve balance between the accuracy of the trained model and the risk of over-fitting. It should be noted that in the original dataset, each feature is a time series of $1000$ points, which is computationally demanding for training the network. In our study, we downsample the data to $100$ points per feature by taking one sample every $10$ points. After downsampling, one training can be finished within $3$ minutes using a PC with Intel(R) Core(TM) i7-8850H CPU, $64$ GB of RAM and an Nvidia Quadro P1000 GPU. 

\section{Results and discussions}\label{Sect_5}

\subsection{Training process}

\begin{figure*}[htb]
    \centering
    \includegraphics[width=.65\textwidth]{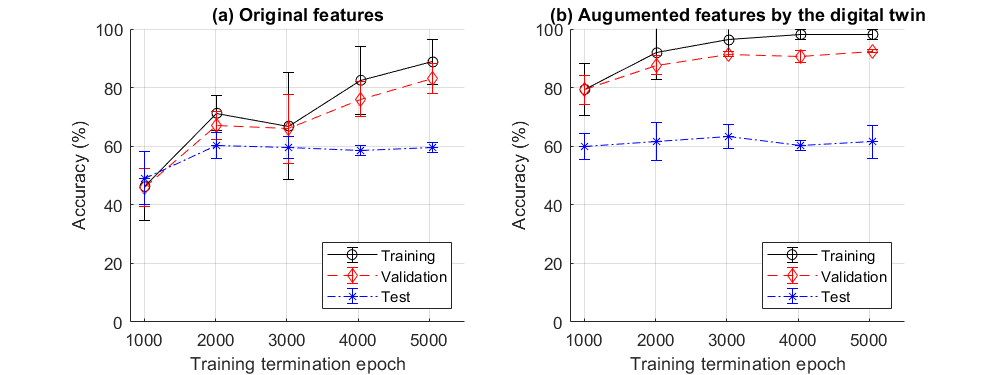}
    \caption{Training process with different termination epochs.}
    \label{fig:trainingValPerfs}
\end{figure*}

Figure \ref{fig:trainingValPerfs} compares the average accuracy over the $9$ classes for the training, validation and test data. The training of the deep learning model is terminated at five different epoch values ranging from $1010$ to $5050$. For each termination epoch value, the experiment is repeated five times and the average accuracy over the five runs as well as the one-standard-deviation error bars (denoted by the two short horizon lines in the Figure) are plotted. The results for the models with original and augmented features are compared in this Figure \ref{fig:trainingValPerfs} (a) and (b). 

From Figure \ref{fig:trainingValPerfs} (a), we can see that after $5050$ epochs, both the training and validation accuracy are still increasing. Therefore, there is still room to improve the model performance on the training and validation dataset if we continue training the model. On the other hand, the test performance has stabilized at around $60\%$ since $2020$ epochs. This indicates that a significant gap might exist on the validation performance and the test performance.

Comparing Figure \ref{fig:trainingValPerfs} (b) to Figure \ref{fig:trainingValPerfs} (a), we can observe that after using the digital twin model to augment the original features, the convergence rate of the training process can be significantly improved. The training and validation accuracy start to stabilize after $3030$ epochs. Also, the variance of the training results are smaller compared to the case of using the original features. The test performance stabilized at around $60\%$ even after $1010$ epochs. After analyzing these two Figures, we decide to terminate the training process after $5050$ epochs to balance the training performance and the risks of overfitting.

\subsection{Average accuracy over all the classes}

In this section, we only analyze the performance of the model with the augmented features as it converges faster compared with the original features and from the initial analysis, there is no significant difference between the test performance of the two models.

Table \ref{table:ResultAcc} shows the average accuracy and its standard deviation for the training, validation and test data. It can be seen that the model, after training for $5050$ epochs, achieved descend accuracy on the training and validation dataset, indicating that the model is capable to diagnose failure accurately in the simulation dataset. The accuracy on the validation dataset is not too much smaller as compared to the training accuracy. This is a positive sign as the model does not have significant problem of overfitting. The mean and standard deviation in Table \ref{table:ResultAcc} are taken over the five repetition of the experiment.

\begin{table}[htb]
    \centering
    \caption{Average accuracy over the $9$ classes.}
    \label{table:ResultAcc}
    \begin{tabular}{ccc}
        \hline
        Dataset & Mean accuracy & Standard deviation \\
        \hline
        Training & $98.12\%$ & 1.71\% \\
        Validation & $92.44\%$ & $0.41\%$ \\
        Test & $61.56\%$ & $5.59\%$ \\
        \hline        
    \end{tabular}
    
\end{table}

\begin{figure*}[htbp]
    \centering
    \includegraphics[width=.8\textwidth, height=.3\textwidth]{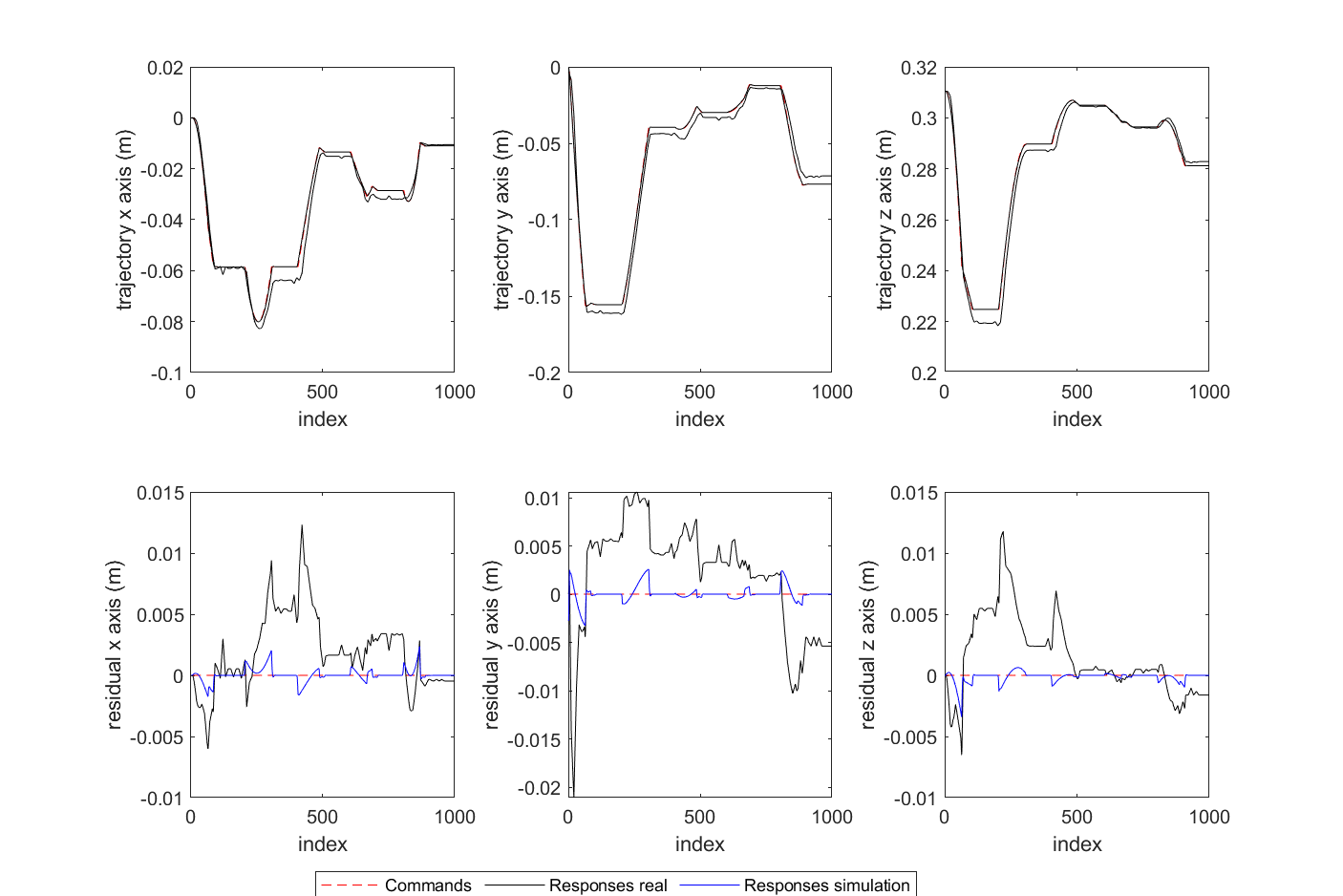}
    \caption{Simulation results from the digital twin model on one test dataset.}
    \label{fig:simulationResultsOnTestDataset}
\end{figure*}

When applied on the test data collected from the real robot, however, the accuracy drops significantly to $61.56\%$. This shows that the fault diagnosis model trained on the simulation data from the digital twin still faces difficulty when applied directly on the real robot. This problem might be caused by various reasons. Among them, the most-likely explanation is that, although the digital twin model has been calibrated with test data, it might not be accurate enough in all the scenarios when the test data are generated. For example, Figure \ref{fig:simulationResultsOnTestDataset} shows the simulation results from the digital twin on one test dataset. The displacement on the $x, y, z$ axes as well as the residual on each axis is shown in this Figure. It can be seen that there is still some discrepancy between the simulation and the real data. Apart from further improving the performance of the digital model, developing methods to train fault diagnosis algorithms that are less sensitive to potential inaccuracy in the simulation model is, then, an important direction of future research.

\subsection{Performances on different classes}

\begin{table*}[htbp]
    \centering
    \caption{Precision, recall and F1 score for each class on the test data.}
    \label{table:otherMetrics}
    \begin{tabular}{ccccc}
        \hline
        index & Class & Precision & Recall & F1 score \\
        \hline
        1 & Healthy & NaN  & $0$ & NaN \\
        2 & Motor 1 steady state error & $1.00\pm0.00$ & $0.52\pm0.13$ & $0.68\pm0.11$ \\
        3 & Motor 1 stuck & $0.96\pm0.06$ & $0.84\pm0.05$ & $0.89\pm0.04$ \\
        4 & Motor 2 steady state error & $0.49\pm0.10$ & $0.70\pm0.10$ & $0.58\pm0.10$ \\
        5 & Motor 2 stuck & $0.75\pm0.10$ & $0.92\pm0.04$ & $0.82\pm0.06$ \\
        6 & Motor 3 steady state error & $0.64\pm0.08$ & $0.70\pm0.16$ & $0.66\pm0.08$ \\
        7 & Motor 3 stuck & $0.64\pm0.16$ & $0.72\pm0.11$ & $0.67\pm0.11$ \\
        8 & Motor 4 steady state error & $0.39\pm0.06$ & $0.68\pm0.11$ & $0.50\pm0.07$ \\
        9 & Motor 4 stuck & $0.49\pm0.11$ & $0.46\pm0.15$ & $0.47\pm0.13$ \\        
        \hline              
    \end{tabular}
\end{table*}

Table \ref{table:otherMetrics} shows the precision, recall and F1 score for each class on the test dataset. As can be seen, class $1$ (no failure) is consistently misclassified by the deep learning model: The model labels all the samples from the healthy class to some other classes. Therefore, the recall for class $1$ is $0$. The precision and F1 score in this case are not defined (NaN) as the model makes no positive predictions. This most likely is caused by the fact that the digital model does not consider the noise and sensor imprecisions presented in the real robot. Therefore, even minor discrepancies present in the real data will make the trained classifier labels them as failure states.

In general, the stuck failures have better performance compared to the steady-state error failures. This is easy to understand as a motor getting stuck has larger impact on the movement of the end-effector than a steady-state error. However, the stuck of motors $3$ and $4$ are difficult to identify, as these two motors are close and it is likely that their stuck make no difference in the observed end-effector motion (see Figure \ref{fig:dtrRobot}).

Another observation is that the failures on motor $1$ (including the stuck and steady-state error) in general have better performance compared to the other motors. This implies that the motor's unique orientation (rotating around the z-axis) causes it's failures to be better identified from the end-effector motion, likely due to their strong impact on the x and y coordinates. The remaining classes are associated to the three motors rotating around the x-axis. They prove generally harder to predict for both methods, often being mistaken one for another. These observations suggest that a possible way to further improve the model performance could be developing dedicated models focusing on the motors with worse performances.

\section{Conclusions and future works}

In this paper, we use digital twins to train deep learning model for component-level fault diagnosis from system-level condition-monitoring data. The digital twin is used for generating data to train an LSTM for fault diagnosis. Further, it is also be used to augmented features used in the LSTM model. The results confirms the potential of using digital twin for training deep learning models for fault diagnosis. It is also found that using a digital twin to augment features can complement the missing component-level information from the system-level condition-monitoring data and improve the performance of the fault diagnosis model.

When applied on the real robot, however, the fault diagnosis model trained by the digital twin still faces difficulty in identifying all the failure modes accurately. One of the main reasons is that the digital twin model inevitably has some simulation errors. We suggest that a transfer Learning approach could be used to improve the performance on real data. Another interesting direction is to try model-based fault diagnosis model where the digital twin is used to simulate the response given different failure mode and the input signals. Then, a fault diagnosis can be easily done by checking the most similar simulation data compared to the measurement. Whilst the performance of the model-based approach might be good, the limitations of this approach are two-fold: first, the significant computational power and time required to simulate the various failure cases; second, the poor adaptability of the method, the performance of which is tied to access to an accurate estimation of the failure's parameters. Because of these limitations, we believe that this approach's main potential is for hybrid methods, to eliminate specific cases (such as lack of failure) with high reliability.

\section*{Acknowledgment}
The research of Zhiguo Zeng is supported by ANR-22-CE10-0004, and chaire of Risk and Resilience of Complex Systems (Chaire EDF, Orange and SNCF).  Killian Mc Court and Xavier Mc Court participate this project as a Pole IA project in the second-year engineering curriculum of Centralesupélec, which are co-supervised by Prof. Zhiguo Zeng and Dr. Lama Itani from Mathworks, and managed by Prof. Wassila Ouerdane from Centralesupélec. The authors would like to also thank the supplier of the robot, Hiwonder for providing technical supports on the robots, and Mr. Xihe Ge from Centralesupélec for procuring of the robot.

{\scriptsize
\bibliography{ref}
\bibliographystyle{ieeetr}
}

\nolinenumbers

\end{document}